# Group Sparse Priors for Covariance Estimation

**Benjamin M. Marlin, Mark Schmidt, and Kevin P. Murphy**

Department of Computer Science
University of British Columbia
Vancouver, Canada

## Abstract

Recently it has become popular to learn sparse Gaussian graphical models (GGMs) by imposing $\ell_1$ or group $\ell_{1,2}$ penalties on the elements of the precision matrix. This penalized likelihood approach results in a tractable convex optimization problem. In this paper, we reinterpret these results as performing MAP estimation under a novel prior which we call the group $\ell_1$ and $\ell_{1,2}$ positive-definite matrix distributions. This enables us to build a hierarchical model in which the $\ell_1$ regularization terms vary depending on which group the entries are assigned to, which in turn allows us to learn block structured sparse GGMs with unknown group assignments. Exact inference in this hierarchical model is intractable, due to the need to compute the normalization constant of these matrix distributions. However, we derive upper bounds on the partition functions, which lets us use fast variational inference (optimizing a lower bound on the joint posterior). We show that on two real world data sets (motion capture and financial data), our method which infers the block structure outperforms a method that uses a fixed block structure, which in turn outperforms baseline methods that ignore block structure.

## 1  Introduction

Reliably estimating a covariance matrix $\Sigma$ is a fundamental problem in statistics and machine learning that arises in many application domains. Covariance estimation is well known to be a statistically challenging problem when the dimensionality of the data $D$ is high relative to the sample size $N$. In the $D > N$ regime, the standard maximum likelihood estimate (the sample covariance matrix $S$) is not positive-definite. Even when $N > D$ the eigenstructure of the sample covariance matrix can be significantly distorted unless $D/N$ is very small (Dempster, 1972).

One particularly promising regularization approach to covariance estimation is to penalize the $\ell_1$-norm of the precision matrix, $\Omega = \Sigma^{-1}$, to encourage sparsity in the precision matrix (Banerjee et al., 2008; Friedman et al., 2007; Yuan and Lin, 2007; Duchi et al., 2008; Schmidt et al., 2009). Zeros in the precision matrix result in absent edges in the corresponding Gaussian graphical model (GGM), so the $\ell_1$-norm can be interpreted as preferring graphs that have few edges. The resulting penalized negative log-likelihood objective function is convex and can be optimized by a variety of methods.

For some kinds of data, it is reasonable to assume that the variables can be clustered (or grouped) into types, which share similar connectivity or correlation patterns. For example, genes can be grouped into pathways, and connections within a pathway might be more likely than connections between pathways. Recent work has extended the above $\ell_1$ penalized likelihood framework to the case of block sparsity by penalizing the $\ell_\infty$-norm (Duchi et al., 2008) or the $\ell_2$-norm (Schmidt et al., 2009) of each block separately; the analogous results for linear regression are known as the simultaneous lasso (Turlach et al., 2005) and the group lasso (Yuan and Lin, 2006) respectively. The resulting objective function is still convex, and encourages block sparsity in the underlying graphs.

For many problems the group structure may not be known a priori, so methods that can simultaneously infer the group structure and estimate the covariance matrix are of great interest. In this paper we convert the $\ell_1$ and group $\ell_{1,2}$ regularization functions into distributions on the space of positive-definite matrices. This enables us to use them as components in larger hierarchical probabilistic models. The key contribution is the derivation of a novel upper bound on the intractable normalizing term that arises in each distribution, which involves an integral over the positive-definite cone. This allows us to lower bound the log-likelihood and derive iterative model fitting procedures that simultaneously estimate the structure and the covariance matrix by optimizing the lower bound. We present analysis and simulation studies investigating properties of the two distributions and their bounds. We apply the bounding framework to the problem of covariance estimation with unknown block structure.

## 2    Related work

The prior distributions we derive and the subsequent covariance estimation algorithms we propose are closely related to the work of Yuan and Lin (2007); Banerjee et al. (2008), which impose sparsity on the elements of the precision matrix $\Omega$ using an $\ell_1$-norm subject to the constraint that $\Omega$ be symmetric and positive-definite. The penalized log-likelihood objective function used to fit the precision matrix is given by

$$\log \det(\Omega) - \text{trace}(\Omega S) - \sum_{i=1}^{D} \sum_{j=1}^{D} \lambda_{ij} |\Omega_{ij}| \qquad (2.1)$$

where $S$ is the sample covariance matrix and $\lambda_{ij} \geq 0$ are the penalty parameters. This objective function is concave for fixed penalty parameters, and various efficient algorithms (typically $O(D^3)$ or $O(D^4)$ time complexity per iteration) have been proposed to solve it (Friedman et al., 2007; Duchi et al., 2008; Schmidt et al., 2009).

In the case where we have known groups of variables, denoted by $G_k$, we can use the following alternative penalization function:

$$- \sum_{kl} \lambda_{kl} \cdot ||\{\Omega_{ij} : i \in G_k, j \in G_l\}||_{p_{kl}} \qquad (2.2)$$

where $p_{kl}$ specifies which norm to apply to the elements from each pair of groups. (Duchi et al., 2008) consider the case $p_{kk} = 1$ within groups and $p_{kl} = \infty$ between groups. (Schmidt et al., 2009) consider the case $p_{kk} = 1$ within groups and $p_{kl} = 2$ between groups, which we refer to as group $\ell_{1,2}$ penalization.

In the limit where each edge is its own group, both penalization functions reduce to the independent $\ell_1$ penalization function. If all diagonal penalty parameters are all equal to some $\lambda > 0$ and the off-diagonal penalty parameters are zero, the optimal precision matrix can be found in closed form by differentiating the penalized log-likelihood objective function. We obtain $\hat{\Omega} = (S + \lambda I)^{-1}$, or equivalently $\hat{\Sigma} = S + \lambda I$. We refer to this method as Tikhonov regularization, following

(Duchi et al., 2008). Tikhonov regularization provides a useful baseline estimator for covariance estimation when $N/D$ is small.

The case of group penalization of GGMs with unknown groups has only been studied very recently. In our previous paper (Marlin and Murphy, 2009), we used a technique that is somewhat similar to the one to be presented in this paper, in that each variable is assigned to a latent cluster, and variables in the same cluster are "allowed" to connect to each other more easily than to variables in other clusters. However, the mechanism by which we inferred the grouping was quite different. Due to the intractability of evaluating the global normalization constant (discussed below), we used *directed* graphical models (specifically, dependency networks), which allowed us to infer the cluster assignment of each variable independently (using EM). Having inferred a (hard) clustering, we then used the penalty in Equation 2.2. In this paper, we instead bound the global normalization constant, and jointly optimize for the group assignments and for the precision matrix parameters at the same time.

We recently came across some independent work (Ambroise et al., 2009) which presents a technique that is very similar to our group $\ell_1$ method (Section 3.2). However, their method has two flaws: First, they ignore the fact that the normalization constant changes when the clustering of the variables changes; and second, they only use local updates to the clustering, which tends to get stuck in local optima very easily (as they themselves remarked). In our paper, we present a mathematically sound derivation of the method, an extension to the group $\ell_{1,2}$ case, and a much better optimization algorithm.

## 3    The Group $\ell_1$ and Group $\ell_{1,2}$ Distributions

It is well known that $\ell_1$ regularized linear regression (i.e., the lasso problem) is equivalent to MAP estimation using a Laplace prior. In this section, we derive the priors that correspond to various $\ell_1$ regularized GGM likelihoods, assuming a known assignment of variables into groups. In later sections, we will use these results to jointly optimize over the precision matrix and the groupings.

### 3.1    The Independent $\ell_1$ Distribution

The penalized log-likelihood in Equation 2.1 (or rather a slight variant in which we only penalize the upper triangle of $X$, since the matrix is constrained to be symmetric) is equivalent to MAP estimation with the following prior, which we call the $\ell_1$ *positive-definite*

*matrix distribution*:

$$P_{L1}(X|\lambda) = \frac{1}{\mathcal{Z}_{I1}} \text{pd}(X) \prod_{i=1}^{D} \prod_{j>i}^{D} \exp(-\lambda_{ij}|X_{ij}|) \quad (3.3)$$

We represent the positive-definiteness constraint using the indicator function pd($X$), which takes the value one if $X$ is positive-definite, and zero otherwise. The normalization term $\mathcal{Z}_{I1}$ is obtained by integrating the unnormalized density over the positive-definite cone. This integral is intractable, but as long as the $\lambda_{ij}$ terms are held fixed, the term is not needed for MAP estimation.

### 3.2   The Group $\ell_1$ Distribution

Our focus in this paper is developing algorithms that infer $\Omega$ under a block structured prior while simultaneously estimating the blocks. The $\ell_1$ distribution does not have a block structure by default, so we augment it with an additional layer of discrete group indicator variables. We assume that the data variables are partitioned into $K$ groups, and we let $z_i$ indicate the group membership of variable $i$. We encourage group sparsity by constraining the $\lambda_{ij}$ such that $\lambda_{ij} = \lambda_0$ if $i$ and $j$ are in different blocks, and $\lambda_{ij} = \lambda_1$ if $i$ and $j$ are in the same block, where $\lambda_0 > \lambda_1$. In addition, we introduce a separate parameter for the diagonal of the precision matrix $\lambda_{ii} = \lambda_D$. The full distribution is given below (where we use the Kronecker delta notation $\delta_{i,j} = 1$ if $i = j$ and $\delta_{ij} = 0$ otherwise).

$$P_{GL1}(X|\lambda, \boldsymbol{z}) = \frac{1}{\mathcal{Z}_1} \text{pd}(X) \prod_{i}^{D} \exp(-\lambda_D|X_{ii}|)$$

$$\times \prod_{i=1}^{D} \prod_{j>i}^{D} \exp(-(\lambda_1 \delta_{z_i,z_j} + \lambda_0(1 - \delta_{z_i,z_j}))|X_{ij}|) \quad (3.4)$$

We refer to this as the *group $\ell_1$ positive-definite matrix distribution*. Note that, despite the name, this distribution does not enforce group sparsity as such, since each element is independently penalized. However, elements in the same group share the same regularizer, which will encourage them to "behave" similarly in term of their sparsity pattern. (This is equivalent to the model considered in Ambroise et al. (2009).)

### 3.3   The Group $\ell_{1,2}$ Distribution

We now derive a prior which yields behavior equivalent to the penalty term in Equation 2.2 for the case $p_{kl} = 2$ (again, we only penalize the upper triangle of the matrix). Unlike the previous group $\ell_1$ prior, this group $\ell_{1,2}$ prior has the property that elements within the same group are penalized together as a group. More precisely, under the group $\ell_{1,2}$ regularization function the precision entry between a pair of variables $\{i,j\}$ within the same group is penalized using an $\ell_1$-norm with penalty parameter $\lambda_1$, except for diagonal entries which are penalized with $\lambda_{ii} = \lambda_D$. For each pair of distinct groups $k, l$, the between group precision entries are penalized jointly under an $\ell_2$-norm with a penalty parameter equal to $\lambda_{kl} = C_{kl} \lambda_0$ where $C_{kl}$ is the product of the size of group $k$ and the size of group $l$ (Schmidt et al., 2009). Scaling by the group size (or any power of the group size greater than 1/2) ensures that the between-group penalties are always greater than the within-group penalties when $\lambda_0 > \lambda_1$. The corresponding prior is shown below; we refer to it as the *group $\ell_{1,2}$ positive-definite matrix distribution*.

$$P_{GL1L2}(X|\lambda, \boldsymbol{z}) = \frac{1}{\mathcal{Z}_{12}} \text{pd}(X) \prod_{i=1}^{D} \exp(-\lambda_D|X_{ii}|)$$

$$\times \prod_{i=1}^{D} \prod_{j>i}^{D} \exp(-\lambda_1 \delta_{z_i,z_j}|X_{ij}|)$$

$$\times \prod_{k=1}^{K} \prod_{l>k}^{K} \exp\left(-\lambda_0 C_{kl} \left(\sum_{i=1}^{D} \sum_{j=1}^{D} \delta_{z_i,k} \delta_{z_j,l}(X_{ij})^2\right)^{1/2}\right)$$

$$\quad (3.5)$$

### 3.4   Group Sparsity Property

The main property of the group $\ell_1$ and group $\ell_{1,2}$ distributions that we are interested in is the suppression of matrix entries between groups of variables. To investigate this property we develop Gibbs samplers for both the group $\ell_1$ and group $\ell_{1,2}$ distributions (see Appendix A for details of the samplers). We consider the illustrative case $D = 4$ yielding five distinct partitions (groupings) of the variables. We run each Gibbs sampler on each grouping with the settings $\lambda_D = \lambda_1 = 0.1$ and $\lambda_0 = 1$. We record a sample after each complete update of the matrix $X$. We collect a total of 1000 samples for each grouping.

In Figure 1 we show estimates of the expected absolute values of $X_{ij}$ ($E[|X_{ij}|]$) for the group $\ell_1$ distribution (the figure is nearly identical in the group $\ell_{1,2}$ case). Studying the expected absolute value of $X_{ij}$ is necessary to reveal the structure in $X$ that results from the underlying partition of the data variables, since off-diagonal terms can be positive or negative. As can clearly be seen in Figure 1, the off-diagonal terms between groups are suppressed while off diagonal terms within groups are not. This is exactly what we would expect based on the group structure of the distribution.

An interesting and unanticipated result is that larger groups appear to have larger diagonal entries under both distributions. Based on the mean of an independent exponential distribution, a reasonable hypothesis for the current parameters would be average diagonal entries of $1/0.1 = 10$ units, which is the case for the partition where every variable is in its own group. However, the partition with all variables in the same group $[1, 1, 1, 1]$ shows a significantly higher diagonal of approximately 15 units. This result clearly illustrates coupling between entries in the matrix $X$ induced by the positive-definite constraint.

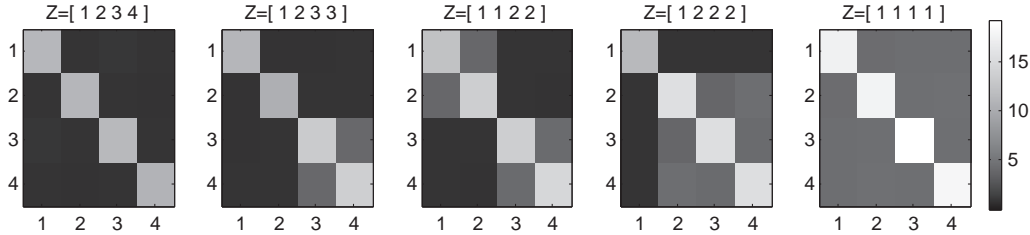

Figure 1: This figure shows an estimate of the mean of the absolute value of $X$ ($E[|X|]$) under the group $\ell_1$ distribution in four dimensions. The parameters used are $\lambda_D = \lambda_1 = 0.1$ and $\lambda_0 = 1$. Estimates under the group $\ell_{1,2}$ distribution are nearly identical to the group $\ell_1$ case shown here.

# 4   Lower Bounds

We recall that for fixed structure and penalty parameters, the estimation of the precision matrix $\Omega$ under either the group $\ell_1$ or group $\ell_{1,2}$ prior distribution is easy since the normalizing term in each distribution is independent of $\Omega$. The difficulty lies in updating the group structure since the intractable normalization term is not constant with respect to changes in the assignment of variables to groups. In this section we derive upper bounds on the intractable normalization terms that allow us to lower bound the log posterior.

## 4.1   Group $\ell_1$ Bound

We first note that the unnormalized densities are always non-negative. As a result, increasing the volume of the domain of integration when computing the normalization term will provide an upper bound on the intractable integral. Instead of integrating over the positive-definite cone $\mathbb{S}_{++}^D$, we integrate over the strictly larger space of symmetric matrices with a positive diagonal. We denote this space of matrices by $\mathbb{S}_P^D$. In the case of the group $\ell_1$ distribution, the integrand completely decouples into independent parts corresponding to standard univariate Laplace and exponential integrals, yielding the following analytic solution for the bound.

$$\mathcal{Z}_1 \leq \int_{\mathbb{S}_P^D} \prod_{i=1}^{D} \prod_{j \geq i}^{D} \exp(-\lambda_{ij}|X_{ij}|) dX \quad (4.6)$$

$$= \prod_{i=1}^{D} \frac{1}{\lambda_{ii}} \cdot \prod_{i=1}^{D} \prod_{j>i}^{D} \frac{2}{\delta_{z_i,z_j} \lambda_1 + (1 - \delta_{z_i,z_j}) \lambda_0} \quad (4.7)$$

## 4.2   Group $\ell_{1,2}$ Bound

We now derive an upper bound on the normalization term for the group $\ell_{1,2}$ distribution. We again apply the strategy of increasing the volume of the domain of integration from the set of positive-definite matrices to the set of symmetric matrices with positive diagonal.

$$\mathcal{Z}_{12} \leq \int_{\mathbb{S}_P^D} \prod_{i=1}^{D} \exp(-\lambda_D |X_{ii}|) \prod_{i=1}^{D} \prod_{j>i}^{D} \exp(-\lambda_1 \delta_{z_i,z_j} |X_{ij}|)$$

$$\cdot \prod_{k=1}^{K} \prod_{l>k}^{K} \exp\left(-\lambda_0 C_{kl} \left(\sum_{i=1}^{D} \sum_{j=1}^{D} \delta_{z_i,k} \delta_{z_j,l} (X_{ij})^2\right)^{1/2}\right) dX \quad (4.8)$$

Unlike the group $\ell_1$ case, the between-block precision entries are coupled together by the $\ell_2$-norm so the integrand does not completely decouple across the upper triangular portion of the matrix $X$. However, a convenient expression for the bound can still be obtained by breaking up the integral into a product of diagonal, within-group, and between-group terms. We introduce the auxiliary variables $C_T = \sum_i \sum_{j>i} \delta_{z_i,z_j}$ to represent the total number of within-group entries across all blocks, and $C_{kl}$ to represent the number of precision entries between variables in group $k$ and group $l$.

$$\mathcal{Z}_{12} \leq \prod_{i=1}^{D} \int_{0}^{\infty} \exp(-\lambda_D |x|) dx \cdot \prod_{i=1}^{C_T} \int_{-\infty}^{\infty} \exp(-\lambda_1 |x|) dx$$

$$\cdot \prod_{k,l \neq k}^{K} \int_{\mathbb{R}^{C_{kl}}} \exp\left(-\lambda_0 C_{kl} \left(\sum_{i=1}^{C_{kl}} x_i^2\right)^{1/2}\right) d\boldsymbol{x} \quad (4.9)$$

The solution to the first two integrals in the bound are standard univariate exponential and Laplace normalization constants. The integral resulting from the between block entries is the normalization constant for the multivariate Laplace distribution in $C_{kl}$ dimensions (Gómez et al., 1998). This allows us to complete the bound as follows.

$$\mathcal{Z}_{12} \leq \left(\frac{1}{\lambda_D}\right)^D \left(\frac{2}{\lambda_1}\right)^{C_T} \prod_{k,l \neq k}^{K} \frac{\pi^{\frac{C_{kl}-1}{2}} \Gamma\left(\frac{C_{kl}+1}{2}\right) 2^{C_{kl}}}{(\lambda_0 C_{kl})^{C_{kl}}}$$

(4.10)

### 4.3 Evaluation of the Bounds

In the case where $D = 2$ it is possible to obtain the exact normalizing constant for the group $\ell_1$ distribution for arbitrary values of $\lambda_D = \lambda_1$ and $\lambda_0$ and the two possible clusterings $z_1 = z_2$ and $z_1 \neq z_2$. We note that in two dimensions the group $\ell_1$ and group $\ell_{1,2}$ distributions are equivalent, $\mathcal{Z}_{12} = \mathcal{Z}_1 = \mathcal{Z}$. To obtain the normalizing term we evaluate the following integral where $\lambda_{12} = \lambda_1$ if $z_1 = z_2$ and $\lambda_{12} = \lambda_0$ is $z_1 \neq z_2$.

$$\mathcal{Z} = \int_{\mathbb{S}_{++}^D} \exp(-\lambda_1(|X_{11}| + |X_{22}|)) \exp(-\lambda_{12}|X_{12}|) dX$$

$$= \int_0^\infty \int_0^\infty \int_{-\sqrt{ab}}^{\sqrt{ab}} \exp(-\lambda_1(a+b) - \lambda_{12}|c|) da\,db\,dc$$

(4.11)

In the case $z_1 = z_2$ the integral evaluates to $(8\pi\sqrt{3} - 18)/(27\lambda_1^3)$. We note that the value of the bound in this case is $2/\lambda_1^3$. The bound thus overestimates the true normalizing term by a constant multiplicative factor approximately equal to 2.115, corresponding to an overestimation of the log normalization term by a constant additive factor equal to 0.7491. In the case $z_1 \neq z_2$ we have obtained an explicit formula for the normalizing term that is defined everywhere except $2\lambda_1 = \lambda_0$. The solution is significantly more complex as seen in Equation 4.12. We have verified empirically that the function is positive and real valued except at the noted singularity.

$$\mathcal{Z} = -\frac{\lambda_0}{2\left(\lambda_1^2 - \frac{1}{4}\lambda_0^2\right)\lambda_1^2} + \frac{\arctan\left(2\frac{\sqrt{\lambda_1^2 - \frac{1}{4}\lambda_0^2}}{\lambda_0}\right)}{\left(\lambda_1^2 - \frac{1}{4}\lambda_0^2\right)^{-3/2}}$$

(4.12)

In Figure 2(a) we show a plot of the error under the bound and under a Monte-Carlo approximation as a function of $\lambda_1$ and $\lambda_0$ in the two dimensional case. We show results for the two unique partitions [1, 1] and [1, 2]. We use a simple importance sampling method for the Monte-Carlo estimate where the proposal distribution is Wishart with a fixed scale matrix equal to the identity matrix and 2 degrees of freedom. We draw 100,000 samples. The primary trend in the error of the bound is revealed by plotting the error as a function of $\lambda_0/\lambda_1$. The error rapidly and smoothly decreases as a function of $\lambda_0/\lambda_1$. The reason for this is that as $\lambda_0/\lambda_1$ increases, the support of the group $\ell_1$ distribution collapses onto the sub-space of diagonal matrices where the bound is exact. The Monte-Carlo estimate of the log normalization term has approximately zero error over the whole range of $\lambda_0/\lambda_1$ values for both partitions.

We extend our analysis to the four dimensional case in Figure 3 where we plot the estimated error between the log bound and log normalizing term for each of the five partitions as a function of $\lambda_1$ and $\lambda_0$. We use the Wishart importance sampler to estimate the normalization terms with scale matrix equal to the identity matrix and 4 degrees of freedom. We draw $10^7$ samples. Similar to the exact analysis in the two dimensional case, we see that the minimum discrepancy between the bound and the true normalizing term occurs for the case where all data dimensions are in their own groups and $\lambda_0/\lambda_1$ is large. The largest discrepancies occur in the case where all data dimensions are in the same group and $\lambda_0 = \lambda_1$.

In Figure 2(b) we show a plot of the bound and the Monte-Carlo estimate of the normalizing term as a function of the matrix dimension $D$. We use $\lambda_1 = 0.1$ and $\lambda_0 = 0.5$. We consider the partition where every dimension is in the same group (1 grp), and the partition where every dimension is in its own group ($D$ grps). Initial investigations suggested that the bound is tightest for the $D$ groups case and weakest for the one group case, so we only consider these two partitions. We also note again that the group $\ell_1$ and group $\ell_{1,2}$ distributions are equivalent for these two partitions. The results show that bound in the one group case diverges more rapidly from the corresponding Monte-Carlo estimate of the true log normalizing term compared to the bound on the $D$ groups case. The fact that the discrepancy in the bound changes as a function of the grouping is somewhat troubling as it may bias model selection towards models with more groups.

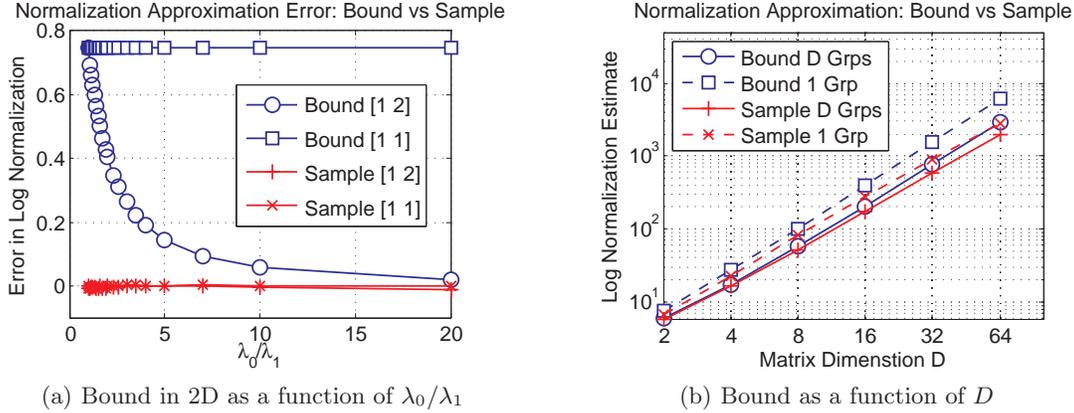

(a) Bound in 2D as a function of $\lambda_0/\lambda_1$      (b) Bound as a function of $D$

Figure 2: Figure (a) shows the approximation error for the log normalization term under the bound and the Monte-Carlo estimate as a function of $\lambda_0/\lambda_1$ in two dimensions. Figure (b) shows both the bound and Monte-Carlo estimate of the log normalization term as the number of dimensions $D$ is varied for $\lambda_0 = 0.5$ and $\lambda_1 = 0.1$.

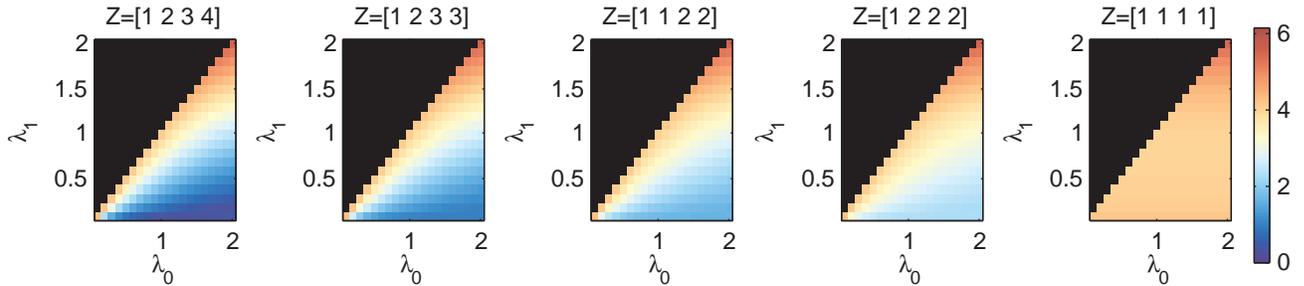

Figure 3: This figure shows the approximation error for the log normalization term under the group $\ell_1$ bound as a function of $\lambda_1$ and $\lambda_0$ for each of the five partitions in four dimensions. This figure is best viewed in color.

## 5 Block Sparse Precision Estimation With Unknown Blocks

In this section we describe a hierarchical block-structured model for precision estimation using the group $\ell_1$ and group $\ell_{1,2}$ prior distributions, and discuss strategies for fitting the models. The hierarchical model includes a Gaussian likelihood term, a discrete distribution over the group indicators $z_i$ with parameter $\theta$, and a symmetric Dirichlet prior distribution on $\theta$ with parameter $\alpha_0/K$ where $K$ is the number of groups. The prior distribution on the precision matrix $P_G(\Omega|\lambda, \boldsymbol{z})$ can be either the group $\ell_1$ distribution $P_{GL1}(\Omega|\lambda, \boldsymbol{z})$ or the group $\ell_{1,2}$ distribution $P_{GL12}(\Omega|\lambda, \boldsymbol{z})$.

$$P(\theta|\alpha_0) = \mathcal{D}(\theta; \alpha_0) \tag{5.14}$$

$$P(z_i = k|\theta) = \theta_k \tag{5.15}$$

$$P(\Omega|\lambda, \boldsymbol{z}) = P_G(\Omega|\lambda, \boldsymbol{z}) \tag{5.16}$$

$$P(\boldsymbol{x}_n|\mu, \Omega) = \mathcal{N}(\boldsymbol{x}_n; \mu, \Omega^{-1}) \tag{5.17}$$

Plugging the upper bound for the normalization term of the group $\ell_1$ or group $\ell_{1,2}$ distribution into the com-plete data log posterior yields an initial lower bound. In addition we employ a variational Bayes approximation $q(\theta|\alpha)$ for the posterior on the mixing proportions $\theta$, where $q(\theta|\alpha)$ is a Dirichlet distribution. This is necessary since the size of $\theta$ varies according to the number of groups, so such parameters need to be integrated out to perform proper model comparison. In the group $\ell_1$ case we employ a fully factorized variational distribution on the group indicator variables $q(z_i = k|\phi_i)$, where $\phi_i$ is a discrete distribution over the $K$ groups. The variational Bayes approximations further lower bound the log posterior. In the group $\ell_{1,2}$ case, the mixture indicators are coupled through the bound on the normalization term, so we work directly with the discrete group indicator variables. We show the bound on the log posterior for the group $\ell_1$ case, using the variational Bayes approximation, in Equation 5.14. (The notation $\tilde{p}(\Omega|z)$ refers to the unnormalized distribution, and $\tilde{\mathcal{Z}}_1$ is our approximate bound from Equation 4.7. The last line corresponds to the entropy of the variational distribution.) The derivation of the bound for the group $\ell_{1,2}$ case is very similar.

$$\log p(\Omega|X) = \log \int \sum_z p(X|\Omega)p(\Omega|z)p(z|\theta)p(\theta|\alpha_0)d\theta \geq \log \int \sum_z p(X|\Omega)\frac{1}{\hat{\mathcal{Z}}_1(z)}\tilde{p}(\Omega|z)p(z|\theta)p(\theta|\alpha_0)d\theta \quad (5.13)$$

$$= \sum_z -\log \hat{\mathcal{Z}}_1(z) + [\log p(X|\Omega) + \log \tilde{p}(\Omega|z)] + \log \int [p(z|\theta)p(\theta|\alpha_0)d\theta]$$

$$\geq \max_{\alpha,\phi} E_q \left\{ -\log \hat{\mathcal{Z}}_1(z) + [\log p(X|\Omega) + \log \tilde{p}(\Omega|z)] + \log p(z|\theta) + \log p(\theta|\alpha_0) - \log q(z,\theta|\alpha,\phi) \right\}$$

$$\geq \max_{\alpha,\phi,\Omega} \frac{N}{2}(-\log(2\pi) + \log\det(\Omega)) - \frac{N}{2}\text{trace}(\Omega S) + \log(pd(\Omega)) + \sum_{i=1}^{D}(-\lambda_D|\Omega_{ii}| + \log(\lambda_D))$$

$$+ \sum_{i=1}^{D}\sum_{j>i}^{D}(-\log(2) + E_q[\delta_{z_i,z_j}](-\lambda_1|\Omega_{ij}| + \log(\lambda_1)) + (1 - E_q[\delta_{z_i,z_j}])(-\lambda_0|\Omega_{ij}| + \log(\lambda_0))$$

$$+ \sum_{i=1}^{D}\sum_{k=1}^{K}E_q[\delta_{z_i,k}]E_q[\log(\theta_k)] + \log(\Gamma(\alpha_0)) - K\log(\Gamma(\alpha_0/K)) + \sum_{k=1}^{K}(\alpha_0/K - 1)E_q[\log\theta_k]$$

$$- \sum_{i=1}^{D}\sum_{k=1}^{K}E_q[\log(\phi_{ik})] - \log(\Gamma(\sum_{k=1}^{K}\alpha_k)) + \sum_{k=1}^{K}\log(\Gamma(\alpha_k)) - \sum_{k=1}^{K}(\alpha_k - 1)E_q[\log\theta_k]$$

In the group $\ell_1$ case, model estimation consists of optimizing $\Omega$, and the variational parameters $\alpha$ and $\phi$. In the group $\ell_{1,2}$ case, model estimation consists of optimizing $\Omega$, the variational parameters $\alpha$, and the partition $z$. The strategy we use for both prior distributions is to start with all the data dimensions in the same group. On each iteration we propose splitting one group into two sub-groups. Given the updated partition $z$ (or updated variational parameters $\phi$) we employ the convex optimization procedure developed by (Duchi et al., 2008) to update the precision matrix under the group $\ell_1$ prior, and the convex optimization procedure developed by (Schmidt et al., 2009) to update the precision matrix under the group $\ell_{1,2}$ prior. The computational cost of model estimation is dominated by the precision matrix updates, which are themselves iterative with an $O(D^3)$ cost per iteration. Finally, we evaluate the lower bound on the log likelihood for the new grouping and parameter settings. We accept the split if it results in an increase in the lower bound. We then update the variational $\alpha$ parameters, which have the simple closed-form update $\alpha_k = \alpha_0 + \sum_{i=1}^{D}\phi_{ik}$ in the group $\ell_1$ case and $\alpha_k = \alpha_0 + \sum_{i=1}^{D}\delta_{z_i,k}$ in the group $\ell_{1,2}$ case. In the group $\ell_1$ case we update the variational $\phi_{ik}$ parameters, which also have a simple closed-form solution. In the group $\ell_{1,2}$ case we perform a local update for each group indicator $z_i$ by reassigning it to the group that gives the maximum value of the bound on the posterior. Each of these steps is guaranteed to increase the value of the bound, and we continue splitting clusters until no split is found that increases the bound.

The key to making the algorithm efficient is the choice of split proposals. We propose a split for a given group by running a graph cut algorithm on a weighted graph derived from the current precision matrix. More precisely, let $U = \{i : z_i = k\}$ be the set of variables belonging to group $k$, and $\overline{U}$ be the other variables. In the group $\ell_1$ case we use the MAP assignments under the variational posterior $z_i = \max_k \phi_{ik}$. We propose a split by computing a normalized cut of the weighted graph $W = |\Omega(U,U)| + 0.5|\Omega(U,\overline{U})||\Omega(U,\overline{U})^T|$, which measures the similarity of variables within group $k$ to each other, as well as the similarity in their relationships to other variables.

We consider two different methods for choosing which groups to split. In the first method, we compute the optimal split for each group. We sort the groups in ascending order according to the weight of the cut divided by the number of variables in the group. We evaluate the split for each group by updating all the model parameters given the new group structure. We accept the first split that results in an increase in the bound on the log posterior. If none of the splits are accepted, we terminate the model estimation procedure. We refer to this as the *greedy method*. In the second method, we exhaustively evaluate the split for all groups. To save on computation time we perform an approximate update for the precision matrix where we only update precision entries between each variable in the group we are splitting and all of the other variables. This is a substantial savings when the groups become small. We select the split giving the highest value of the bound, perform a full update on the precision matrix, and re-compute the bound on the log posterior. If the selected split fails to increase the bound on the log posterior, we terminate the model estimation procedure. We refer to this as the *exhaustive method*.

## 6    Covariance Estimation Experiments

In this section we apply the group $\ell_1$ and group $\ell_{1,2}$ distributions to the regularized covariance estimation problem. We consider the group $\ell_1$ greedy method for unknown groups (GL1-ug), the group $\ell_1$ exhaustive method for unknown groups (GL1-ue), the group $\ell_{1,2}$ greedy method for unknown groups (L12-ug), and the group $\ell_{1,2}$ exhaustive method for unknown groups (GL12-ue). We compare against three other methods: Tikhonov regularization (T), independent $\ell_1$ regularization (IL1), and group $\ell_{1,2}$ regularization with known groups (GL12-k). We compute test set log likelihood estimates using five-fold cross validation. We hold out an additional one fifth of the training set to use as a validation set for selecting the penalization parameters $\lambda_D, \lambda_1, \lambda_0$. For each penalty parameter we consider 10 values from $10^4$ to 1 equally spaced on a log scale. We consider all combinations of values subject to the constraint that $\lambda_0 > \lambda_1 > 0.5\lambda_D$. We set $\alpha_0 = 1$. We center and scale all of the data before estimating the models. We report test set log likelihood results using the parameters that achieve the maximum average validation log likelihood.

### 6.1    CMU Motion Capture Data Set

In this section, we consider the motion-capture data set used in our previous work (Marlin and Murphy, 2009). This consists of 100 data cases, each of which is 60 dimensional, corresponding to the $(x, y, z)$ locations of 20 body markers. These were manually partitioned into five parts (head and neck, left arm, right arm, left leg, and right leg), which we refer to as the known structure.

We give test log likelihood results for the CMU data set in Figure 6a. All of the methods that estimate the group structure from the data (GL1-ug, GL1-ue, GL12-ug, GL12-ue) significantly out perform the unstructured Tikhonov (T) and independent $\ell_1$ methods (IL1), and give an improvement over the group $\ell_{1,2}$ method with known groups based on body parts (GL12-k). The best method on the CMU data set is the group $\ell_1$ exhaustive search method (GL1-ue). Figure 6b gives the total training time (based on a Matlab implementation) over all crossvalidation folds and parameter settings. Note that the vertical axis is on a log scale. We can see that all of the methods that estimate the group structure require significantly more computation time relative to the unstructured Tikhonov and independent $\ell_1$ methods, as well as the group

$\ell_{1,2}$ method with known groups (GL12-k). However, the computation times among the methods that estimate the group structure are all quite similar.

We show the known or inferred group structure for each of the group methods in Figures 5a-e. We show results for the fold with the highest test log likelihood. All of the methods that estimate group structure appear to over-partition the data variables relative to the known structure. However, the over-partitioning is quite systematic and mostly corresponds to breaking up the given groups into sub-groups corresponding to their x-coordinates, y-coordinates, and z-coordinates (note that the ordering of the variables is $x_1, y_1, z_1, x_2, y_2, z_2, ...$). The apparent over-partitioning also results in improved test log likelihood relative to the known groups, indicating that it is well supported by the data, and not simply an artifact of the bounds.

### 6.2    Mutual-Fund Data Set

The second data set we consider consists of monthly returns for 59 mutual-funds in four different sectors including 13 US bond funds, 30 US stock funds, 7 balanced funds investing in both $US$ stocks and bonds and 9 international stock funds. There are 86 data cases each corresponding to the returns of all 59 funds for a different month. While the funds are naturally split into groups based on their content, the groups are clearly not independent since the balanced funds group contains both stocks and bonds. This data set has been used previously by Scott and Carvalho (2008) in the context of local search for decomposable GGM graph structure.

We give test log likelihood results for the mutual-funds data set in Figure 6c. We first note that there is much less variation in median test log likelihood across the methods compared to the CMU data set, which likely results from the mutual fund data set having a less obvious block structure. Indeed, the group $\ell_{1,2}$ method (GL12-k) based on the fund-type grouping yields a median test set log likelihood that is only slightly better than the independent $\ell_1$ method. All of the methods that estimate the group structure from the data (GL1-ug, GL1-ue, GL12-ug, GL12-ue) result in median test log likelihood performance that is no worse than the independent $\ell_1$ method. The best method overall is again the group $L1$ method with exhaustive search (GL1-ue). This method in fact yields better performance than the independent $\ell_1$ method across all test folds. The trend in the computation time results is very similar to the CMU data set and is not shown.

We present the known or inferred structure for each of the group methods in Figures 5f-j. We show results for the fold with the highest test log-likelihood. Unlike the CMU case, some of the unknown group methods select many more groups than the known grouping, while others select less. The group $\ell_1$ exhaustive search method (GL1-ue), which obtains the best test log likelihood, selects 12 groups. Interestingly, it recovers the bonds group with only one error, splits the international stock funds into two groups, but mixes the balanced funds with the US stock funds in several small groups. The group $\ell_{1,2}$ exhaustive search method (GL12-ue) apparently terminates after correctly splitting the variables into one group of bond funds and one group of all other funds. The group $\ell_{1,2}$ greedy search method (GL12-ug) recovers the bonds and international stock groups with only one error while mixing the US stocks and balanced funds into two groups.

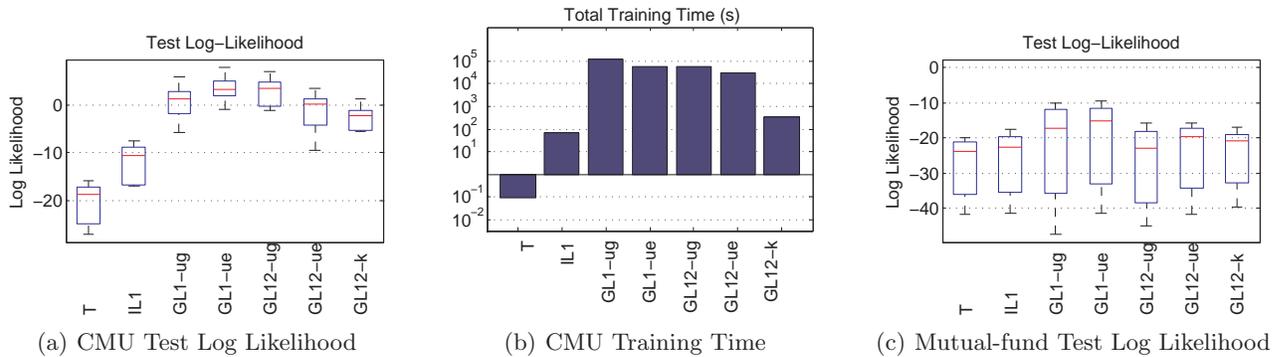

Figure 4: Test set log likelihood and training time for CMU, and test log likelihood for and mutual-funds.

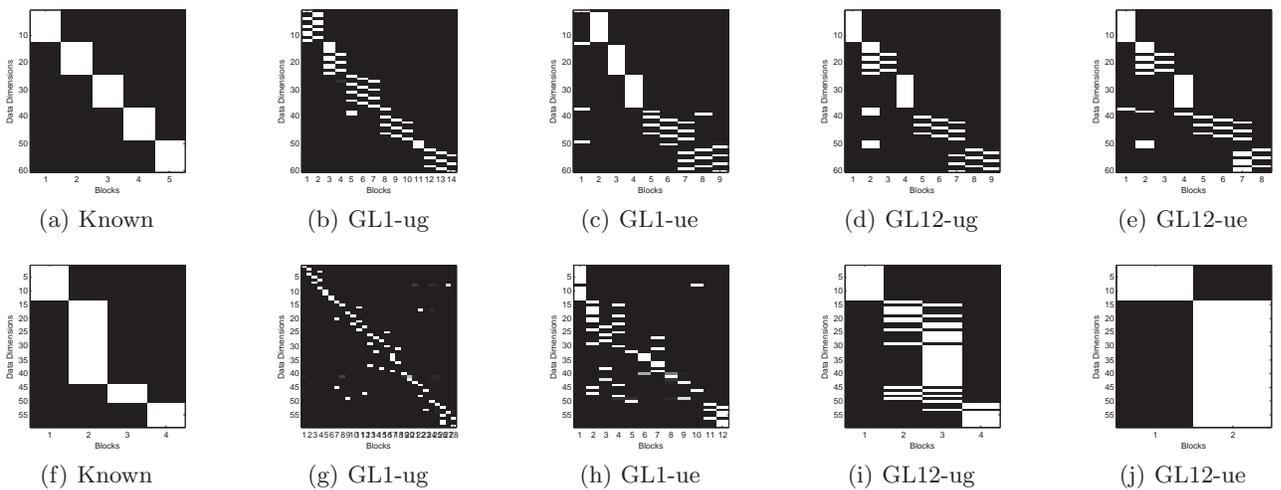

Figure 5: Inferred clusterings on the CMU motion capture data (top row) and mutual funds data (bottom row).

## A   Gibbs Sampler Outline

Suppose that $P(X)$ is an arbitrary density function defined over the space of positive-definite matrices. To implement a Gibbs sampler we require the conditional distribution $P(X_{ij}|X_{-ij})$ where $X_{-ij}$ denotes all the entries in $X$ except for $X_{ij}$ and $X_{ji}$. Independent of the form of the density, the positive-definiteness constraint implies that the conditional distribution for $X_{ij}$ will only have support on an interval $b_0(X_{-ij}) < X_{ij} < b_1(X_{-ij})$. Due to the fact that the positive-definite cone is a convex set, this interval, which is the intersection of a line with the positive-definite cone, will also be convex.

The exact end points $b_0$ and $b_1$ can be obtained in closed form for any $i, j$ and matrix $X$. We omit the full derivation due to space limitations, but sketch a brief outline. First, we note that $b_0$ and $b_1$ are the maximum and minimum values for $X_{ij}$ that render $X$ indefinite. Finding them reduces to the problem of solving the equation $\det(X) = 0$ in terms of $X_{ij}$. Assuming $X$ is otherwise positive-definite, the determinant is a linear function of a diagonal entry $X_{ii}$, leading to a finite positive lower bound $b_0$ and an upper bound $b_1 = \infty$. The determinant is a non-degenerate quadratic function of an off diagonal entry $X_{ij}$ leading to finite values for $b_0$ and $b_1$. These results are intuitive since increasing a diagonal entry of a matrix that is already positive-definite will keep it positive-definite, while sufficiently increasing or decreasing an off diagonal entry will make it violate positive-definiteness.

Assuming we have derived the allowable range for $X_{ij}$ given $X_{-ij}$ we consider particular cases for the density function $P(X)$. In the case of the group matrix $\ell_1$ distribution we obtain the following conditional distribution, which is easily seen to be a truncated Laplace distribution for off diagonal entries, and a truncated exponential distribution for on diagonal entries given the results for $b_0$ and $b_1$ that we have just derived. Sampling from these truncated distribution is simple using inversion of the corresponding cumulative distribution functions.

$$P(X_{ij}|X_{-ij}) = \frac{(X_{ij} \in [b_0, b1]) \exp(-\lambda_{ij}|X_{ij}|)}{\int_{b_0}^{b_1} \exp(-\lambda_{ij}|X_{ij}|)dX_{ij}}$$

The group matrix $\ell_{1,2}$ distribution has identical conditional distributions for diagonal and off-diagonal-within-block entries. The off diagonal between block entries have the form of a truncated hyperbolic distribution due to the application of the $\ell_2$-norm. We give the result below assuming that dimension $i$ belongs to group $k$ and dimension $j$ belongs to group $l$. Sampling from the truncated hyperbolic distribution can also be done efficiently by exploiting the fact that this form of the hyperbolic distribution is a generalized inverse gaussian (GIG) scale mixture with zero mean. We can sample the scale parameter from the correct GIG distribution, and then use inversion of the CDF to sample from a truncated univariate normal distribution with the sampled scale parameter.

$$P(X_{ij}|X_{-ij}) = \frac{(X_{ij} \in [b_0, b1]) \exp\left(-\lambda_{kl}\sqrt{\gamma_{ij}^2 + X_{ij}^2}\right)}{\int_{b_0}^{b_1} \exp\left(-\lambda_{kl}\sqrt{\gamma_{ij}^2 + X_{ij}^2}\right)dX_{ij}}$$

$$\gamma_{ij} = \left(\sum_{s \neq i}\sum_{t \neq j, t > s}^{D} \delta_{z_s,k}\delta_{z_t,l}X_{st}^2\right)^{1/2}$$

The complexity of the sampler is dominated by the calculation of the the truncation range $[b_0, b_1]$. Solving for $b_0$ and $b_1$ requires inverting a matrix of size $D-1$, at a cost of $O((D-1)^3)$. The complete cost of $T$ updates to all the unique matrix parameters is $O(\frac{1}{2}TD(D+1)(D-1)^3)$, or approximately $O(TD^5)$, which is clearly intractable unless $D$ is small.